\documentclass{article}

\usepackage{microtype}
\usepackage{graphicx}
\usepackage{subcaption}
\usepackage{booktabs} 

\usepackage{hyperref}

\usepackage[preprint]{icml2026}

\usepackage{amsmath}
\DeclareMathOperator*{\argmin}{arg\,min}
\usepackage{amssymb}
\usepackage{mathtools}
\usepackage{amsthm}

\usepackage[capitalize,noabbrev]{cleveref}

\theoremstyle{plain}

\theoremstyle{definition}

\theoremstyle{remark}

\usepackage[textsize=tiny]{todonotes}

\icmltitlerunning{DreamPRM-Code}

\begin{document}

\twocolumn[
  \icmltitle{DreamPRM-Code: Function-as-Step Process Reward Model with Label Correction for LLM Coding}

  \icmlsetsymbol{equal}{*}

  \begin{icmlauthorlist}
    \icmlauthor{Ruiyi Zhang}{ucsd}
    \icmlauthor{Peijia Qin}{ucsd}
    \icmlauthor{Qi Cao}{ucsd}
    \icmlauthor{Pengtao Xie}{ucsd}
  \end{icmlauthorlist}

  \icmlaffiliation{ucsd}{University of California, San Diego}
  \icmlcorrespondingauthor{Pengtao Xie}{p1xie@ucsd.edu}

  \icmlkeywords{Machine Learning, ICML}

  \vskip 0.3in
  
]

\printAffiliationsAndNotice{}  

\begin{abstract}
Process Reward Models (PRMs) have become essential for improving Large Language Models (LLMs) via test-time scaling, yet their effectiveness in coding remains limited due to the lack of meaningful step decompositions in code and the noise of Monte-Carlo–generated partial labels. We propose DreamPRM-Code, a coding-focused PRM that treats functions as reasoning steps using a Chain-of-Function prompting strategy to induce modular code generation, enabling PRM training and application analogous to mathematical reasoning tasks. To address label noise, DreamPRM-Code introduces a meta-learning–based correction mechanism that leverages clean final-solution unit-test labels and performs bi-level optimization to refine intermediate labels. Applying on test-time scaling, DreamPRM-Code achieved state-of-the-art performance on LiveCodeBench with 80.9 pass@1 rate, surpassing OpenAI o4-mini. \\

Project Page: \href{http://github.com/ruz048/DreamPRM-Code}{[DreamPRM-Code]} 
\end{abstract}

\section{Introduction } Post-training methods for Large Language Models (LLMs)~\citep{brown2020language,qiu2025gated}, such as Reinforcement Learning from Human Feedback (RLHF)~\citep{ouyang2022training} and test-time scaling~\citep{openai2024o1,guo2025deepseekr1}, have proven effective for improving model capability. In this process, Process Reward Models (PRMs), which take in a partial reasoning state of an LLM and output a correctness score, have emerged as a promising method~\citep{wang-etal-2024-multi-step,zhang2024restmcts,cao2025dreamprm}. They are capable of (1) simulating costly human-feedback rewards to conduct reinforcement learning for LLMs~\citep{zou2025reasonfluxprm} and (2) supporting test-time scaling of LLMs via PRM-selected best-of-N or tree search methods~\citep{setlur2025rewarding}.

Although PRMs have seen widespread success in mathematical reasoning tasks, their application to coding tasks remains limited. While solving both coding and mathematical problems requires strong reasoning capability from LLMs, two challenges prevent PRMs from being effectively used for coding. First, most modern PRMs require LLM-generated text to be separated into meaningful steps. Unlike Chain-of-Thought (CoT) reasoning in mathematics~\citep{wei2022chain}, which naturally decomposes into steps, LLM-generated code lacks a clear definition of such ``steps.'' Some existing works use a line of code as a reasoning step, which significantly increases computation cost due to too many steps~\citep{wang-etal-2024-multi-step}. Other define steps in natural-language-based planning processes instead of in the generated code, ignoring inherent reasoning process in programming code itself~\citep{li-etal-2025-codeprm}. Second, PRMs are typically trained with Monte-Carlo-generated pseudo-labels for the correctness of each partial reasoning state. These labels can be quite noisy and ultimately reduce PRM performance. 

To tackle these challenges, we propose a novel method called DreamPRM-Code. To tackle the first challenge, the proposed method leverages a novel Chain-of-Function prompting strategy to use functions as PRM steps. To address the second challenge, DreamPRM-Code contains a novel label correction method to automatically purify noisy middle-step PRM labels via meta learning. We benchmark DreamPRM-Code on ~\citep{jain2025livecodebench} and achieved 80.9 pass@1 rate, surpassing the strongest existing model, OpenAI o4-mini~\citep{openai_o3_o4mini_system_card}, and other baseline test-time scaling methods.

\section{Method} 

In this section, we describe the proposed DreamPRM-Code framework in detail. DreamPRM-Code is a coding-focused Process Reward Model that defines reasoning steps at the level of functions in the generated code and incorporates a label-correction mechanism to improve training signal quality.

\subsection{Chain-of-Function as PRM Steps}
To address the first challenge outlined in the previous section, namely the absence of a natural step decomposition in code, we introduce a Chain-of-Function prompting strategy designed to encourage LLMs to generate modular, well-structured programs. This prompt steers the model toward producing independent code blocks whose logic can be isolated and encapsulated into separate functions. Inspired by Chain-of-Thought prompting, where high-level plans are produced before detailed reasoning, Chain-of-Function encourages the model to generate high-level functions early, providing a blueprint that guides the construction of more specific helper functions. Under this structure, DreamPRM-Code can treat each function as a coherent reasoning step, enabling the direct adaptation of standard PRM training and inference pipelines commonly used in mathematical reasoning tasks.

Algorithm~\ref{alg} illustrates an example decomposition under the Chain-of-Function paradigm, where a model-generated solution is separated into three PRM steps. In the first step, a \texttt{main} function is introduced to select Dijkstra’s algorithm as the appropriate approach for the problem. The remaining two steps implement the \texttt{dijkstra} and \texttt{build\_graph} functions, which are invoked within \texttt{main} to complete the overall solution. This structured decomposition not only provides clear intermediate reasoning states for PRM evaluation but also aligns closely with human software-engineering practices, enabling more interpretable and modular code generation.

\begin{algorithm}[htbp]
\caption{An Example of Chain-of-Function Definition for PRM Steps in LLM Generated Code}
\begin{algorithmic}[]
\label{alg}
\STATE
\STATE \hspace{1cm} \texttt{Chain-of-Function Prompt}

\STATE
\STATE \quad You will be given a question (problem specification) and will generate a correct Python program that matches the specification and passes all tests. You should follow the code organization guidelines:
\STATE
\STATE \quad {1. Function Organization:} Place high-level solving functions first (e.g., ``main()'', primary algorithm). Follow with helper functions that implement specific subtasks. Avoid nested functions; keep all functions at the top level.
\STATE
\STATE \quad {2. Write Docstrings in Each Function (Chain of Thought):} Add a docstring using triple quotes (\texttt{'''}) at the beginning of each function. Main methods should explain the approach, algorithm choice, strategy, and key steps; helper methods should explain their specific logic and purpose.
\STATE
\STATE \quad {3. Extract logic into separate methods} when operations are repeated or complex calculations can be isolated.

\STATE \rule{\linewidth}{0.5pt}
\STATE
\STATE \hspace{2cm} \texttt{LLM Generated Code}
\STATE \textbf{(Step-1)}

\STATE \texttt{def main():}
\STATE \quad \texttt{'''}
\STATE \quad Strategy: Use Dijkstra's algorithm to find shortest path in weighted graph...
\STATE \quad \texttt{'''}
\STATE \quad [Code Block]
\STATE
\STATE \textbf{(Step-2)}

\STATE \texttt{def dijkstra(graph, start, end):}
\STATE \quad \texttt{'''}
\STATE \quad Implements Dijkstra's algorithm using min-heap priority queue...
\STATE \quad \texttt{'''}
\STATE \quad [Code Block]

\STATE \textbf{(Step-3)}

\STATE \texttt{def build\_graph(n, m):}
\STATE \quad \texttt{'''}
\STATE \quad Build adjacency list from stdin input...
\STATE \quad \texttt{'''}
\STATE \quad [Code Block]
\STATE
\end{algorithmic}
\end{algorithm}

\subsection{Automatic Label Correction with Meta Learning}

To address the second challenge, we propose a meta learning based label correction framework~\citep{Tu2023LearningFN,Tao2025Meta,Taraday_2023_ICCV,Wu2020LearningTP} that explicitly denoises MC sampled PRM labels. In coding tasks, we can leverage a private unit test evaluation system to obtain clean and reliable correctness labels once the LLM completes code generation. These final step labels provide strong supervision signals that are not available in mathematical reasoning tasks, where intermediate and final reasoning states typically lack automatic evaluation. DreamPRM-Code exploits this unique property of coding problems to improve the quality of intermediate supervision.

Specifically, we consider two sources of data: clean last step PRM data $(X_{\text{meta}}, Y_{\text{meta}})$ obtained from unit test evaluation, and noisy middle step data $(X, Y)$ obtained through MC sampling. Since the middle step labels may be inaccurate, we treat the noisy labels $Y$ as learnable parameters rather than fixed targets, and optimize them using meta learning. This allows the model to adaptively correct label noise based on performance on clean final step data.

The proposed approach follows a bi-level optimization framework. In the lower level, we train the PRM $f_{\theta}$ on the middle step data using the current estimate of noisy labels. The PRM parameters are updated by minimizing the training loss:

\begin{equation}
    \theta^{*}(Y) = \argmin_{\theta} \mathcal{L}\big(f_{\theta}(X), Y\big),
\end{equation}

where the optimized parameter $\theta^{*}$ depends implicitly on the labels $Y$. In the upper level, we evaluate the optimized PRM $f_{\theta^{*}(Y)}$ on the clean last step data and compute a meta loss with respect to the ground truth labels $Y_{\text{meta}}$:

\begin{equation}
    \min_{Y} \, \mathcal{L}\big(f_{\theta^{*}(Y)}(X_{\text{meta}}), Y_{\text{meta}}\big).
\end{equation}

\begin{figure}
    \centering
    \includegraphics[width=0.8\linewidth]{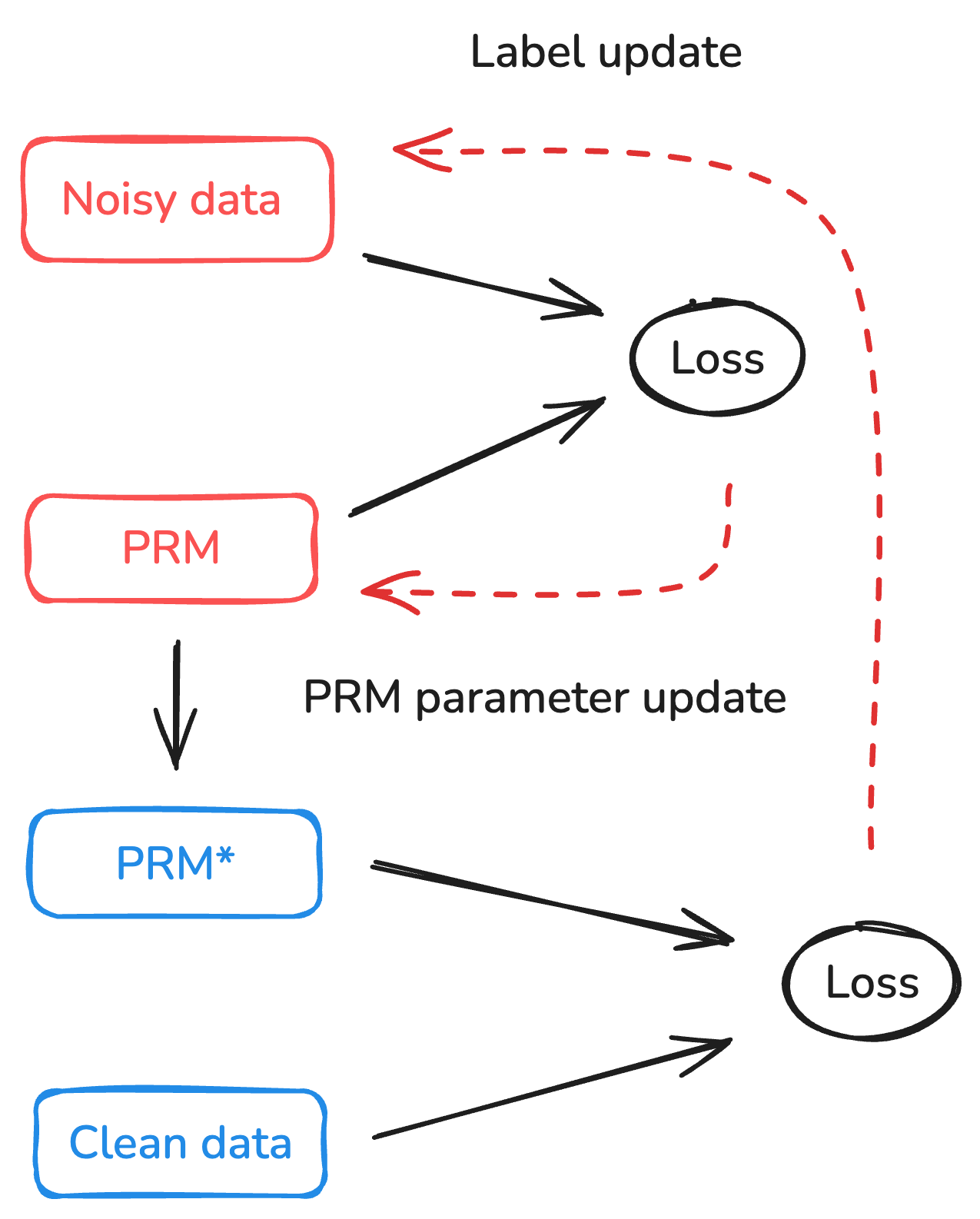}
    \caption{\textbf{The meta-learning based label correction framework in DreamPRM-Code.} The parameters of the PRM are first optimized using noisy data. The updated PRM is then applied on clean data to compute a loss, which is then used for optimize label of the noisy data used in previous step. }
    \label{fig:method}
\end{figure}

Because the meta loss depends on $\theta^{*}$ and therefore on the middle step labels $Y$, minimizing this objective encourages label assignments that lead to PRM parameters generalizing well to clean final step supervision. In practice, we solve this nested optimization by alternately updating the PRM parameters and the noisy labels using gradient descent. Through this iterative process, DreamPRM-Code progressively refines the middle step labels, producing denoised training signals that improve PRM robustness and downstream performance in both test time scaling and reinforcement learning settings. A complete illustration of the label correction process is shown in Figure \ref{fig:method}.

\section{Results} 
\subsection{Experimental Settings}

We evaluate DreamPRM-Code primarily on LiveCodeBench (LCB)~\citep{jain2025livecodebench}, a comprehensive and contamination-free benchmark for code generation with large language models. LiveCodeBench continuously collects newly released programming problems from competitive programming platforms such as LeetCode, AtCoder, and Codeforces. Each problem is associated with a timestamp indicating its publication date. In our experiments, we use 601 problems published before 2024-08-01 as the training split for PRM, and 131 problems published after 2025-02-01 as the test split, ensuring a strict temporal separation and no data overlap.

During the training of DreamPRM-Code, we use Qwen3-Coder-30B-A3B~\citep{Yang2025Qwen3TR} to generate both Chain-of-Function (CoF) trajectories for PRM training and Monte Carlo (MC)–sampled labels. We adopt Qwen-2.5-Coder-3B as the process reward model (PRM)\citep{Yang2024Qwen25TR}. Specifically, we replace the final language-model head with a classification head to output step-level reward scores. We train the PRM using the Adam optimizer~\citep{Kingma2014AdamAM} with a learning rate of $10^{-4}$, a meta-learning rate of $10^{-2}$, and a weight decay of $10^{-2}$. We use a batch size of 8 and set the meta-learning unroll step to 1. The bi-level optimization procedure for label correction is implemented using the Betty library\citep{choe2023betty}.

For test-time scaling, we use OpenAI o4-mini-high as the base (policy) model~\citep{openai_o3_o4mini_system_card}. For each problem, the base LLM generates four candidate solutions, from which DreamPRM-Code selects the final output. We apply mean aggregation over the per-step reward scores produced by DreamPRM-Code to obtain a final score for each candidate.

\subsection{Baselines}

We compare DreamPRM-Code against two categories of baselines. First, we consider leading LLMs on LiveCodeBench without test-time scaling, including o4-mini-high~\citep{openai_o3_o4mini_system_card}, Gemini-2.5~\citep{Comanici2025Gemini2P}, DeepSeek R1~\citep{guo2025deepseekr1}, and the o3 models.

Second, we evaluate alternative test-time scaling approaches, including outcome reward models (ORMs), which assign a single reward score to the entire generated solution rather than per-step rewards~\citep{Uesato2022SolvingMW}. We also include an ablation baseline of DreamPRM-Code without label correction (PRM-CoF).

\subsection{Results on LiveCodeBench}
\begin{table}[t]
\centering
\caption{Performance comparison of DreamPRM-Code and baseline methods across Easy, Medium, and Hard difficulty levels on LiveCodeBench. Results include top-performing LLMs from the leaderboard and test-time scaling methods applied to O4-mini-high, with results reported as pass@1 percentages.}
\begin{tabular}{lcccc}
\toprule
 & Easy & Medium & Hard & Overall \\
\midrule
\multicolumn{5}{c}{\textit{Top LLMs on the Leaderboard}} \\
\midrule
Gemini-2.5 & \bf 100 & 82.1 & 52.5 & 72.5 \\
O3 & \bf 100 & 71.8 & 57.4 & 71.8 \\
DeepSeek-R1 & 99.7 & 77.7 & 47.2 & 68.7 \\
O4-mini-high & \bf 100 & 89.7 & 57.4 & 77.1 \\
\midrule
\multicolumn{5}{c}{\textit{Test-Time Scaling Methods (O4-mini-high)}} \\
\midrule
ORM & \bf 100 & 89.7 & 62.3 & 79.4 \\
\bf DreamPRM-Code & \bf 100 & \bf 92.3 & \bf 63.9 & \bf 80.9 \\
\bottomrule
\end{tabular}
\label{tab:results}
\end{table}

The results of DreamPRM-Code and all baselines are reported in Table~\ref{tab:results}. Compared to leading proprietary models on the LiveCodeBench leaderboard, DreamPRM-Code outperforms the strongest baseline, o4-mini-high, by 3.8 points in pass@1 accuracy. We further compare DreamPRM-Code with other test-time scaling methods under the same base model, o4-mini-high. DreamPRM-Code consistently outperforms ORM-based scaling, highlighting the advantage of decomposing code generation into intermediate steps and assigning rewards at a finer granularity, and underscoring the importance of our Chain-of-Function (CoF) formulation. In summary, the state-of-the-art performance of DreamPRM-Code demonstrates the effectiveness of our proposed approach.

\subsection{Ablation Study}

\begin{table}[t]
\centering
\caption{Performance comparison between DreamPRM-Code and its ablation baseline, DreamPRM-Code without the label correction mechanism (PRM-CoF).}
\begin{tabular}{lcccc}
\toprule
 & Easy & Medium & Hard & Overall \\
\midrule
PRM-CoF & \bf 100 & \bf 92.3 & 62.3 & 80.2 \\
\bf DreamPRM-Code & \bf 100 & \bf 92.3 & \bf 63.9 & \bf 80.9 \\
\bottomrule
\end{tabular}
\label{tab:abl}
\end{table}

In this section, we conduct an ablation study to investigate the effectiveness of the label correction mechanism in DreamPRM-Code by comparing it with an ablation baseline in which PRMs are trained on CoF trajectories using traditional MC-sampled labels without label correction. As shown in Table \ref{tab:abl}, DreamPRM-Code outperforms this ablation baseline (PRM-CoF), demonstrating the effectiveness of the proposed label correction framework.

\section{Conclusion}
In this paper, we propose DreamPRM-Code, a process reward model designed for large language model–based code generation. DreamPRM-Code introduces a novel Chain-of-Function prompting strategy to explicitly define and generate fine-grained PRM steps, along with a meta-learning–based label correction approach to mitigate noise in intermediate-step training labels. When applied to test-time scaling, DreamPRM-Code achieves state-of-the-art performance on LiveCodeBench, attaining an 80.9 pass@1 score and outperforming OpenAI o4-mini as well as other test-time scaling baselines. These results demonstrate the effectiveness of DreamPRM-Code in improving step-level reasoning quality and overall code generation performance.

\bibliography{example_paper}
\bibliographystyle{icml2026}

\newpage
\appendix
\onecolumn


\end{document}